\documentclass[sigconf]{acmart}

\AtBeginDocument{%
  \providecommand\BibTeX{{%
    \normalfont B\kern-0.5em{\scshape i\kern-0.25em b}\kern-0.8em\TeX}}}

\usepackage{microtype}
\usepackage{graphicx}
\usepackage{subfigure}
\usepackage{booktabs} 

\usepackage{hyperref}
\usepackage{multirow}
\usepackage{xcolor,colortbl}

\usepackage{stackengine,scalerel}

\newcommand\obullet[1]{\ThisStyle{\ensurestackMath{%
  \stackon[1pt]{\SavedStyle#1}{\SavedStyle\kern.6\LMpt\bullet}}}}
\newcommand\ocirc[1]{\ThisStyle{\ensurestackMath{%
  \stackon[1pt]{\SavedStyle#1}{\SavedStyle\kern.6\LMpt\circ}}}}
 
\usepackage{amsmath}
\usepackage{mathtools}
\usepackage{amsthm}
\usepackage{mathtools}
\usepackage{wrapfig}
\usepackage[capitalize,noabbrev]{cleveref}
\usepackage{algorithmic}
\usepackage[linesnumbered,ruled,noend]{algorithm2e} 
\usepackage{diagbox}
\usepackage{booktabs}
\usepackage{hhline}

\usepackage{balance}
\newcommand{\std}{\scriptsize{}}

\copyrightyear{2024} 
\acmYear{2024} 
\setcopyright{acmlicensed}\acmConference[WSDM '24]{Proceedings of the 17th ACM International Conference on Web Search and Data Mining}{March 4--8, 2024}{Merida, Mexico}
\acmBooktitle{Proceedings of the 17th ACM International Conference on Web Search and Data Mining (WSDM '24), March 4--8, 2024, Merida, Mexico}
\acmPrice{15.00}
\acmDOI{10.1145/3616855.3635831}
\acmISBN{979-8-4007-0371-3/24/03}

\begin{document}

\title{Continuous-time Autoencoders for Regular and Irregular Time Series Imputation}


\author[1]{Hyowon Wi}
\affiliation{%
  \institution{Yonsei University}
  \streetaddress{50 Yonsei-ro}
  \city{Seoul}
  \country{South Korea}}
\email{wihyowon@yonsei.ac.kr}

\author[2]{Yehjin Shin}
\affiliation{%
  \institution{Yonsei University}
  \streetaddress{50 Yonsei-ro}
  \city{Seoul}
  \country{South Korea}}
\email{yehjin.shin@yonsei.ac.kr}

\author[3]{Noseong Park}
\affiliation{%
  \institution{Yonsei University}
  \streetaddress{50 Yonsei-ro}
  \city{Seoul}
  \country{South Korea}}
\email{noseong@yonsei.ac.kr}
\authornote{Corresponding author.}



\renewcommand{\shortauthors}{Wi et al.}

\begin{abstract}
Time series imputation is one of the most fundamental tasks for time series. Real-world time series datasets are frequently incomplete (or irregular with missing observations), in which case imputation is strongly required. Many different time series imputation methods have been proposed. Recent self-attention-based methods show the state-of-the-art imputation performance. However, it has been overlooked for a long time to design an imputation method based on \emph{continuous-time} recurrent neural networks (RNNs), i.e., neural controlled differential equations (NCDEs). To this end, we redesign time series (variational) autoencoders based on NCDEs. Our method, called \emph{continuous-time autoencoder} (CTA), encodes an input time series sample into a continuous hidden path (rather than a hidden vector) and decodes it to reconstruct and impute the input. In our experiments with 4 datasets and 19 baselines, our method shows the best imputation performance in almost all cases.
\end{abstract}



\maketitle

\section{Introduction}

Time series is one of the most frequently occurring data formats in real-world applications, and there exist many machine learning tasks related to time series, ranging from stock price forecasting to weather forecasting~\cite{khare2017short,vargas2017deep,dingli2017financial,jiang2021applications,sen2021accurate,hwang2021climate,choi2023NADE,karevan2020transductive,hewage2020temporal,yu2017spatio,wu2019graph,fang2021spatial,li2021spatial,tekin2021spatio, zeng2022transformers, zhou2021informer, shi2015convolutional}. These applications frequently assume complete time series. In reality, however, time series can be incomplete with missing observations, e.g., a weather station's sensors are damaged for a while. As a matter of fact, many famous benchmark datasets for time series forecasting/classification are pre-processed with imputation methods to make them complete, e.g.,~\cite{chen2001freeway,JIANG2022117921,choi2022STGNCDE, choi2023stgnrde}. In this regard, time series imputation is one of the most fundamental topics in the field of time series processing. However, its difficulty lies in that i) time series is incomplete since some elements are missing and moreover, ii) the missing rate can be sometimes high, e.g.,~\cite{silva2012predicting}.
\begin{figure}[!t]
    \begin{subfigure}[Latent ODE~\cite{latentode} produces a hidden vector $\mathbf{z}(0)$ given a time series input. We note that the decoder of Latent ODE reverts the sequence of the input, and only one pair of $(\boldsymbol{\mu},\boldsymbol{\sigma})$ is produced. The incomplete observation in red is first pre-processed by setting missing elements to zeros.]{\includegraphics[width=0.90\columnwidth]{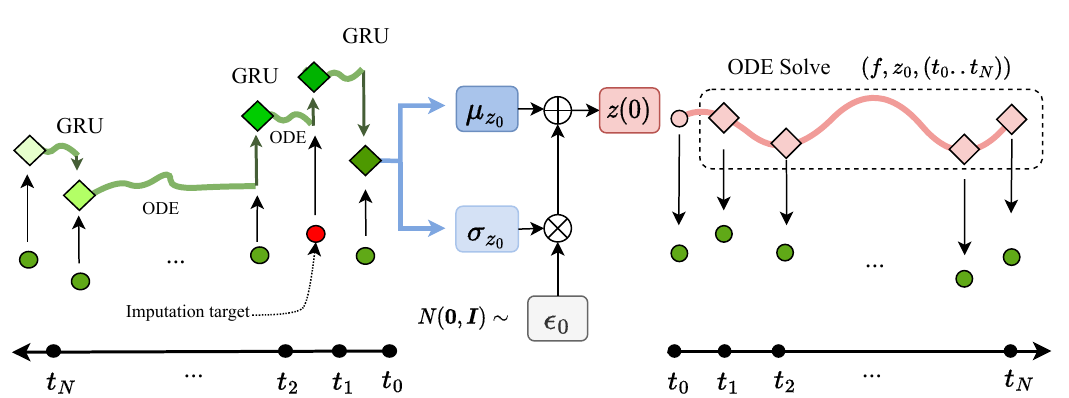}
        \label{fig:latentODE}}
    \end{subfigure}
    \begin{subfigure}[Our proposed CTA, whose encoder/decoder are NCDEs, produces a continuous hidden path $H(t), t \in {[}t_0, t_N{]}$. Given a discrete time series sample, the encoder first interpolates it to estimate its continuous data path (highlighted in light green) and then evolves its state vector, i.e., ${[}\boldsymbol{\mu}(t), \boldsymbol{\sigma}(t){]}$ in this example, via the Riemann–Stieltjes integral. The decoder reconstructs and imputes for the original incomplete time series sample. Since ${[}\boldsymbol{\mu}(t), \boldsymbol{\sigma}(t){]}$ is defined continuously over time, one can consider that there exist \emph{infinitely many} autoencoders between $t_0$ and $t_N$ (cf. Fig.\ref{fig:infinite_autoencoder}).]{\includegraphics[width=0.90\columnwidth]{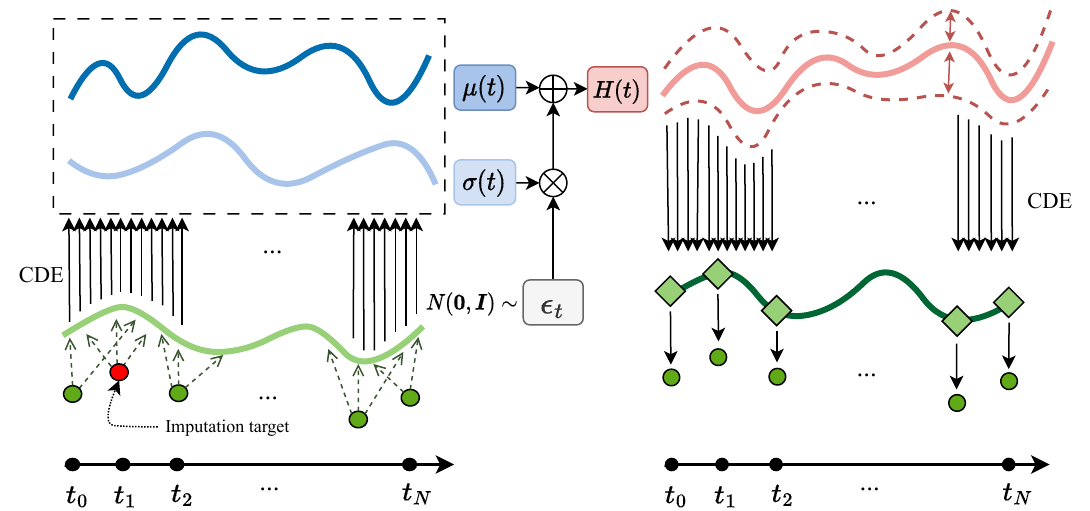}
        \label{fig:CTA}}
    \end{subfigure}
   \caption{Overall workflow of Latent ODE and our method, CTA, to impute incomplete observations in red.}    
\vspace{-1em}
\end{figure}

\begin{table}[!t]
\caption{Comparison of methods in terms of i) how to process irregular time series, and ii) how to process the missing value of the original data, and iii) whether the time can be handled continuously}
\label{table:intro}
\begin{center}
\setlength{\tabcolsep}{2pt}
\begin{small}
\begin{tabular}{l|c|c|c}
\specialrule{1pt}{1pt}{1pt}
Model & irregular time series & missing value & continuous time \\
\specialrule{0.5pt}{0.5pt}{0.5pt}
BRITS~\cite{brits} & time decay & \multirow{4}{*}{fill zero}  & \multirow{4}{*}{X} \\
GP-VAE~\cite{gp-vae} & raw timestamp & &  \\
GAIN~\cite{yoon2018gain} & X & &  \\
SAITS~\cite{saits} & positional encoding & &  \\
\specialrule{0.5pt}{0.5pt}{0.5pt}
CTA  & \multicolumn{2}{c|}{neural controlled differential equation} & O \\
\specialrule{1pt}{1pt}{1pt}
\end{tabular}
\end{small}
\end{center}
\vspace{-1em}
\end{table}

To this end, diverse approaches have been proposed, ranging from simple interpolations to deep learning-based methods. Those deep learning-based methods can be further categorized into recurrent neural network-based~\cite{suo2019recurrent,mrnn,gru-d,brits}, variational autoencoder-based~\cite{latentode,gp-vae,hi-vae}, generative adversarial networks-based~\cite{yoon2018gain,luo2018multivariate,luo2019e2gan}, self-attention-based~\cite{cdsa,shan2021nrtsi,shukla2021mtan,shukla2021hetvae,saits}, and some others. Among them, self-attention-based methods, e.g., SAITS~\cite{saits}, show the state-of-the-art imputation quality. SAITS adopts dual layers of transformers since the time series imputation is challenging and therefore, a single layer of transformer is not sufficient.

\paragraph{\textbf{Our approach:}} In this work, we propose the concept of \underline{\textbf{C}}ontinuous-\underline{\textbf{T}}ime \underline{\textbf{A}}utoencoder (CTA) to impute time series. We, for the first time, extend (variational) autoencoders for processing time series \emph{in a continuous manner} --- there already exist some other \emph{non-continuous-time} autoencoders for time series, e.g., Latent ODE~\cite{latentode}. To enable our concept, we resort to neural controlled differential equations (NCDEs) which are considered as \emph{continuous-time recurrent neural networks} (RNNs)~\cite{hochreiter1997long,cho2014learning}. The overall framework follows the (variational) autoencoder architecture~\cite{kingma2013auto} with an NCDE-based encoder and decoder (see Fig.~\ref{fig:CTA}). Since time series for imputation is inevitably irregular with missing elements, our method based on continuous-time RNNs, i.e., NCDEs, is suitable for the task.

Table~\ref{table:intro} summarizes how the state-of-the-art RNN, VAE, GAN, and transformer-based imputation models process irregular and incomplete time series inputs --- existing methods primarily focus on regular time series and process irregular time series with heuristics, e.g., time decay. BRITS gives a decay on the time lag, and GP-VAE uses the raw timestamp as an additional feature. The temporal information is not used in GAIN, and SAITS adopts the positional encoding within its transformer. Moreover, they fill out missing values simply with zeros, which introduces noises into the data distribution. To address these limitations, our proposed method resorts to the NCDE technology by creating the continuous hidden path with irregular time series inputs. By modeling the hidden dynamics of time series in a continuous manner, in addition, our method is able to learn robust representations. 

\begin{wrapfigure}{r}{0.45\columnwidth}
\vspace{-1.2em}
    \includegraphics[trim={0cm 0.0cm 0cm 0cm},clip, width=0.37\columnwidth]{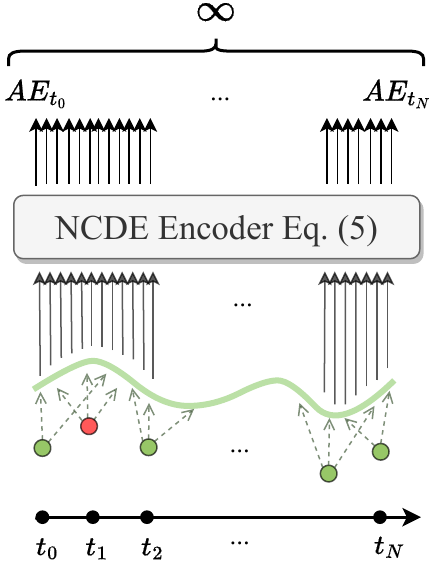}
    \caption{We highlight the encoder part from Fig.~\ref{fig:CTA}.}
    \label{fig:infinite_autoencoder}

\end{wrapfigure}

\paragraph{\textbf{An infinite number of autoencoders:}} Our approach differs from other (variational) autoencoder-based approaches for time series that encode a time series sample into a single hidden vector and decodes it (cf. Fig.~\ref{fig:latentODE} where $\mathbf{z}(0)$ is produced by the encoder). As shown in Fig.~\ref{fig:CTA}, however, we define an autoencoder for every time $t \in [t_0, t_N]$ given a time series sample. In other words, there exist infinitely many autoencoders in $[t_0, t_N]$ since we can define $[\boldsymbol{\mu}(t), \boldsymbol{\sigma}(t)]$ for every single time point $t$. At the end, the continuous hidden path $H(t)$ is produced (rather than a vector). Therefore, one can consider that our  method is a continuous generalization of (variational) autoencoders --- our method is able to continuously generalize both variational and vanilla autoencoders.

\paragraph{\textbf{Hidden vector vs. hidden path:}} Compared to the single hidden vector approach, our method has much flexibility in encoding an input time series sample. Since the single hidden vector may not be able to compress all the information contained by the input, it may selectively encode some key information only and this task can be difficult sometimes. However, our method encodes the input into a continuous path that has much higher representation flexibility.

\paragraph{\textbf{Dual layer and training with missing values:}} Existing time series imputation methods have various architectures. However, some highly performing methods have dual layers of transformers, e.g., SAITS, and being inspired by them, we also design i) a special architecture for our method and ii) its training algorithm. We carefully connects two continuous-time autoencoders (CTAs) via a learnable weighted sum method, i.e., we learn how to combine those two CTAs. Since our CTA can be either variational or vanilla autoencoder, we test all four combinations of them, i.e., two options for each layer. In general, VAE-AE or AE-AE architectures show good performance, where VAE means variational autoencoder and AE means vanilla autoencoder, and the sequence separated by the hyphen represents the layered architecture. In addition, we train the proposed dual-layered architecture with our proposed special training method with missing elements. We intentionally remove some existing elements to create imputation environments for training.

We conduct time series imputation experiments with 4 datasets and 19 baselines. In almost all cases, our CTA shows the best accuracy and its model size is also much lower than the state-of-the-art baseline. Our contributions can be summarized as follows:
\begin{enumerate}
    \item We generalize (variational) autoencoders in a continuous manner. Therefore, the encoder in our proposed method creates a continuous path $H(t)$ of latent representations, from which our decoder reconstructs the original time series and imputes missing elements. Our continuous hidden path $H(t)$ is able to encode rich information.
    \item We test with various missing rates from 30\% to 70\% on 4 datasets. Our method consistently outperforms baselines in most cases, owing to the continuous RNNs, i.e., NCDEs.
\end{enumerate}

\section{Preliminaries \& Related Work}


\begin{figure}[t]
    \begin{subfigure}[]{\includegraphics[width=0.46\columnwidth]{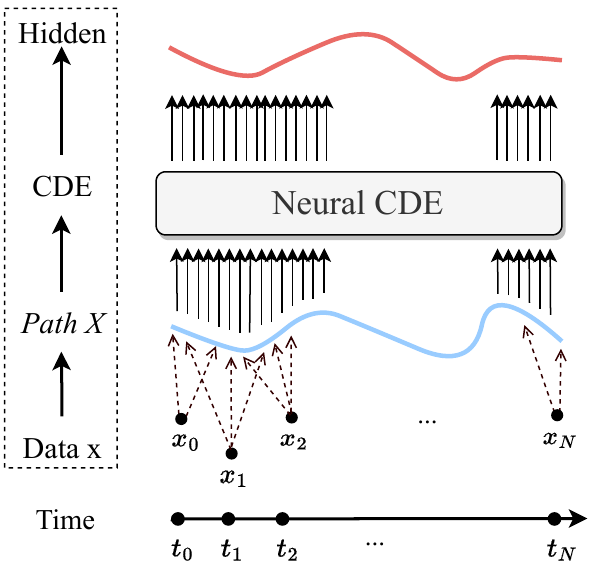}\label{fig:nerualcde}}\end{subfigure}
    \begin{subfigure}[]{\includegraphics[width=0.4\columnwidth]{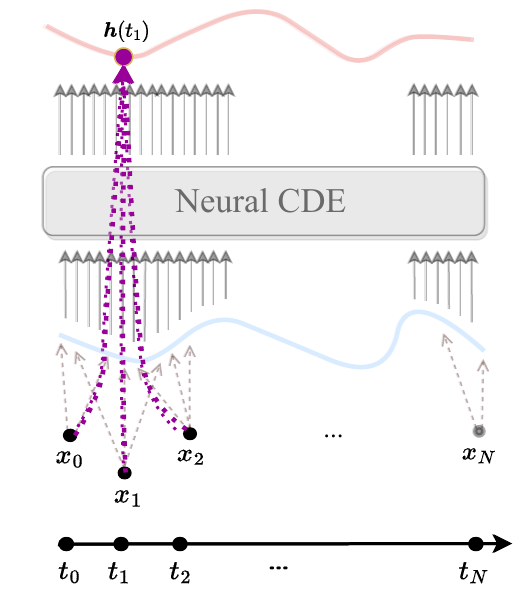}\label{fig:nerualcde2}}\end{subfigure}
    \caption{Overall workflow of NCDEs. In (a), after creating the continuous path $X(t)$, every computation works in the continuous time domain. When calculating $\boldsymbol{h}(t_1)$ in (b), observations around $t_1$ are considered via the interpolated continuous path $X(t)$ --- some sparse transformers also consider a small number of neighboring observations in a discrete time domain~\cite{10.1145/3530811}.}
\vspace{-1em}
    
\end{figure}

\subsection{Neural Controlled Differential Equations}
NCDEs solve the following initial value problem (IVP) based on the Riemann-Stieltjes integral problem~\cite{stieltjes1894recherches} to derive the hidden state $\boldsymbol{h}(T)$ from the initial state $\boldsymbol{h}(0)$:
\begin{align} \label{eq_ode}
    \boldsymbol{h}(T) &= \boldsymbol{h}(0) + \int_0^T f(\boldsymbol{h}(t), t;\boldsymbol{\theta}_f)dX,\\
    &= \boldsymbol{h}(0) + \int_0^T f(\boldsymbol{h}(t), t;\boldsymbol{\theta}_f)\frac{dX(t)}{dt} dt,
\end{align}where $X(t)$ is a path representing the continuous input. In general, $X(t)$ is estimated from its discrete time series $\{(\mathbf{x}_i, t_i)\}_{i=0}^N$ via an interpolation method, where $\mathbf{x}_i$ is a multivariate observation at time $t_i$, e.g., given discrete sensing results of weather stations, we reconstruct their continuous path via an interpolation method --- the natural cubic spline method~\cite{mckinley1998cubic, ncde} is frequently used for NCDEs since it is twice differentiable when calculating the gradient w.r.t. $\boldsymbol{\theta}_f$. For this reason, NCDEs are called as \emph{continuous RNNs} --- one can consider that the hidden state $\boldsymbol{h}(t)$ of RNNs continuously evolves from $t=0$ to $T$ while reading the input $\frac{dX(t)}{dt}$ in NCDEs (cf. Fig.~\ref{fig:nerualcde}). Since NCDEs create $X(t)$ via the interpolation method, however, the hidden state $\boldsymbol{h}(t)$ at time $t$ is created by considering input observations around $t$ (cf. Fig.~\ref{fig:nerualcde2}), which is one subtle but important difference from conventional RNNs.

The above IVP of NCDEs can be solved with existing ODE solvers, since  $\frac{d\boldsymbol{h}(t)}{dt}=f(\boldsymbol{h}(t),t,\theta_f)\frac{dX(t)}{dt}$. For instance the explicit Euler method, the simplest ODE solver, solves the above IVP by iterating the following step multiple times from $t=0$ to $T$:
\begin{align}\label{eq:euler}
\boldsymbol{h}(t + s) &= \boldsymbol{h}(t) + s  \frac{d\boldsymbol{h}(t)}{dt} = \boldsymbol{h}(t) + s  f(\boldsymbol{h}(t),t,\theta_f)\frac{dX(t)}{dt},
\end{align}where $s$ is a pre-configured step size.


\subsection{Time Series Imputation} \label{s:imputation}

\subsubsection{RNN-based}

Existing RNN-based models regard timestamps as one attributes of raw data. GRU-D~\cite{gru-d} proposes the concept of time lag and imputes missing elements with the weighted combination of its last observation and the global mean with time decay. However, such assumption has limitations on general datasets. M-RNN~\cite{mrnn} proposes the multi-directional RNN to impute random missing elements, which considers both intra-data relationships inside a data stream and inter-data relationships across data streams. M-RNN, however, has no consideration on the correlation among features. BRITS~\cite{brits} imputes missing elements with bi-directional RNNs using time decay. It also makes use of bi-directional recurrent dynamics, i.e., they train RNNs in both forward and backward directions, introducing advanced training methods, e.g., a consistency loss function.

\subsubsection{VAE-based}
Latent ODE~\cite{latentode} is VAE-based model that adopts ODE-RNN~\cite{chen2018neural} as its encoder to encode a time series sample to a single hidden vector, and use it as the initial value of its ODE-based decoder. Hence, Latent ODE can handle sparse and/or irregular time series without any assumptions. Sequential VAEs are designed to extent the latent space of VAEs over time, considering the time information of sequential samples~\cite{Xiaoyu2021dvae}. VRNN~\cite{chung2015vrnn} combines VAE and RNN to capture the temporal information of the data. To overcome the deterministic property of RNNs, SRNN~\cite{fraccaro2016srnn} and STORN~\cite{bayer2014storn} propose stochastic sequential VAEs by integrating RNNs and state space models. However, existing sequential VAEs struggle to handle irregular data as they heavily rely on RNNs. GP-VAE~\cite{gp-vae} is sequential VAE-based imputation model which has an assumption that high-dimensional time series has a lower-dimensional representation that evolves smoothly over time using a Gaussian process (GP) prior in the latent space.

\subsubsection{GAN-based}
Recently, generative adversarial networks (GANs) have been used to impute missing values. GAIN~\cite{yoon2018gain} is the first model to apply GANs to the imputation task. The generator replaces missing values based on observed values, while the discriminator determines the correctness of the replaced values compared to the actual values. The discriminator receives partial hints on missing values during training. GRUI-GAN~\cite{luo2018multivariate} is a combination of GRU-D and GAN. It uses the GRU-I structure where the input attenuation is removed. It combines the generator and classifier structures using this modified GRU-I structure to increase accuracy through adversarial learning. E2GAN~\cite{luo2019e2gan} introduces the concept of an end-to-end model. It constructs an autoencoder structure based on GRU-I in the generator. Time series data is compressed into a low-dimensional vector through the autoencoder and used for generation.

\subsubsection{Self-attention-based}

Self-attention mechanisms~\cite{vaswani2017attention} have been adapted for time series imputation after demonstrating an improved performance on seq-to-seq tasks in natural language processing. mTAN~\cite{shukla2021mtan} proposed a model that combines VAEs and multi-time attention module that embeds time information to process irregularly sampled time series. HetVAE~\cite{shukla2021hetvae} can handle the uncertainties of irregularly sampled time series data by adding a module that encodes sparsity information and heterogeneous output uncertainties to the multi-time attention module. SAITS~\cite{saits} uses a weighted combination of two self-attention blocks and a joint-optimization training approach for reconstruction and imputation. SAITS now shows the state-of-the-art imputation accuracy among those self-attention methods. 

\section{Problem Definition}
In many real-world time series applications, incomplete observations can occur for various reasons, e.g., malfunctioning sensors and/or communication devices during a data collection period. As a matter of fact, many benchmark datasets for time series classification and forecasting had been properly imputed before being released~\cite{chen2001freeway,JIANG2022117921,choi2022STGNCDE}. Therefore, imputation is one of the key tasks for time series.

Given a time series sample $\{(\mathbf{x}_i, t_i)\}_{i=0}^N$, where $t_i < t_{i+1}$, and $t_i \in [0,T]$, let $\mathbf{X} \in \mathbb{R}^{\dim(\mathbf{x}) \times N}$ be a matrix-based representation of $\{(\mathbf{x}_i, t_i)\}_{i=0}^N$. We consider real-world scenarios that some elements of $\mathbf{X}$ can be missing. Thus, we denote the incomplete time series with missing elements as $\ddot{\mathbf{X}}$ --- those missing elements can be denoted as \textit{nan} in $\ddot{\mathbf{X}}$. Our goal is to infer $\hat{\mathbf{X}} \approx \mathbf{X}$ from $\ddot{\mathbf{X}}$.

For ease of our discussion but without loss of generality, we assume that i) all elements of $\mathbf{X}$ are known, and ii) $t_0 = 0, t_N = T$. For our experiments, however, some ground-truth elements of $\mathbf{X}$ are missing in its original data, in which case we exclude them from testing and training (see Appendix~\ref{a:missing}).


\section{Proposed Method}
We describe our proposed method in this section. We first outline the overall model architecture, followed by detailed designs.

\subsection{Encoder}

Given an incomplete time series sample $\{(\ddot{\mathbf{x}}_{t_i}, t_i)\}_{i=0}^N$, which basically means $\ddot{\mathbf{X}}$, we first build a continuous path $X(t)$ as in the original NCDE method. We note that after the creation of the continuous path $X(t)$, we have an observation for every $t \in [0,T]$. After that, our NCDE-based encoder begins --- for ease of discussion, we assume variational autoencoders and will shortly explain how they can be changed to vanilla autoencoders. For our continuous-time variational autoencoders, we need to define two continuous functions, $\boldsymbol{\mu}(t): [0,T] \rightarrow \mathcal{R}^{\dim(\boldsymbol{\mu})}$ and $\boldsymbol{\sigma}(t): [0,T] \rightarrow \mathcal{R}^{\dim(\boldsymbol{\sigma})}$, each of which denotes the mean and standard deviation of the hidden representation w.r.t. time $t$, respectively. We first define the following augmented state of $\boldsymbol{e}(t)$, where $\boldsymbol{\mu}(t)$ and $\boldsymbol{\sigma}(t)$ are concatenated into a single vector form:
\begin{align}
    \boldsymbol{e}(t) &= (\boldsymbol{\mu}(t), \boldsymbol{\sigma}(t))
\end{align}

We then define the following NCDE-based continuous-time encoder:
\begin{align}\begin{split}
    \boldsymbol{\mu}(t) &= \boldsymbol{\mu}(0) + \int_0^t g_\mu(\boldsymbol{e}(t), t; \boldsymbol{\theta}_\mu) dX,\\
                &\Rightarrow \boldsymbol{\mu}(0) + \int_0^t g_\mu(\boldsymbol{e}(t), t; \boldsymbol{\theta}_\mu) \frac{dX(t)}{dt} dt, \\
    \boldsymbol{\sigma}(t) &= \boldsymbol{\sigma}(0) + \int_0^t g_\sigma(\boldsymbol{e}(t), t; \boldsymbol{\theta}_\sigma) dX,\\
         &\Rightarrow \boldsymbol{\sigma}(0) + \int_0^t g_\sigma(\boldsymbol{e}(t), t; \boldsymbol{\theta}_\sigma) \frac{dX(t)}{dt} dt, \label{eq1:4}    
\end{split}\end{align}where $\frac{d\boldsymbol{\mu}(t)}{dt}$ and $\frac{d\boldsymbol{\sigma}(t)}{dt}$ are modeled by $g_\mu(\boldsymbol{e}(t), t; \boldsymbol{\theta}_\mu) \frac{dX(t)}{dt}$ and $g_\sigma(\boldsymbol{e}(t), t; \boldsymbol{\theta}_\sigma) \frac{dX(t)}{dt}$, respectively.

The continuous path of the hidden representation of the input time series, which we call as \emph{continuous hidden path} hereinafter, can then be written as follows, aided by the reparameterization trick:

\begin{align}
H(t) &= \boldsymbol{\mu}(t) + \boldsymbol{\epsilon}_t\odot\exp(\boldsymbol{\sigma}(t)),\label{eq1:5}
\end{align}where $\boldsymbol{\epsilon}_t \sim \mathcal{N}(\boldsymbol{0}, \boldsymbol{I})$, and $\odot$ means the element-wise multiplication. 

One subtle point is that we use $\exp(\boldsymbol{\sigma}(t))$ instead of $\boldsymbol{\sigma}(t)$ in Eq.~\eqref{eq1:5}. In other words, $\boldsymbol{\sigma}(t)$ is for modeling the log-variance in our case. In our preliminary study, this log-variance method brings much more stable training processes. The reason is that the exponential function amplifies the continuous log-variance path $\boldsymbol{\sigma}(t)$ and therefore, the continuous variance path by $\exp(\boldsymbol{\sigma}(t))$ can represent complicated sequences. An alternative is to model the continuous variance path directly by $\boldsymbol{\sigma}(t)$, which can be a burden for the encoder.


\paragraph{\textbf{Network architecture:}} Note that $g_\mu$, $g_\sigma$ are neural networks in our method. We basically use fully-connected layers with non-linear activations to build them. 
The architecture of the NCDE functions $g_{\mu}$, $g_{\sigma}$ in the encoder are as follows:
\begin{equation}
   \begin{aligned}
        g_\mu(\boldsymbol{e}(t), t; \boldsymbol{\theta}_\mu) &= \psi(\texttt{FC}(\mathbf{E}_L)) \\
        g_\sigma(\boldsymbol{e}(t), t; \boldsymbol{\theta}_\sigma) &= \psi(\texttt{FC}(\mathbf{E}_L)) \\
        \vdots \\
        \mathbf{E}_1 &= \omega(\texttt{FC}({\mathbf{E}_0})),\\
        \mathbf{E}_0 &= \omega(\texttt{FC}({\mathbf{e}}(t))),
    \end{aligned} 
\end{equation}where $\omega$ is a sigmoid linear unit~\cite{elfwing2018sigmoid}, $\psi$ is a hyperbolic tangent, and $L$ is the number of hidden layers. We use $dim(\mathbf{h})$ to denote the hidden size before the final layer and $dim(\mathbf{l})$ to denote the size of the final layer. Therefore, $\mathbf{E}_i$ has a size of $dim(\mathbf{h})$ for all $i$ and the output sizes of $g_\mu, g_\sigma$ are commonly $dim(\mathbf{l})$.

\subsection{Decoder}
Our NCDE-based decoder, which decodes $H(t)$ into an inferred (or a reconstructed) time series sample, can be written as follows:
\begin{align}\begin{split}
    \boldsymbol{d}(t) &= \boldsymbol{d}(0) + \int_0^T k(\boldsymbol{d}(t); \boldsymbol{\theta}_k) dH,\\
                 &\Rightarrow \boldsymbol{d}(0) + \int_0^T k(\boldsymbol{d}(t); \boldsymbol{\theta}_k) \frac{dH(t)}{dt} dt, \\
                 &\Rightarrow \boldsymbol{d}(0) + \int_0^T k(\boldsymbol{d}(t); \boldsymbol{\theta}_k) \Big(\frac{d\boldsymbol{\mu}(t)}{dt} + \boldsymbol{\epsilon}_t\odot \frac{d\exp(\boldsymbol{\sigma}(t))}{dt}\Big) dt, \label{eq2:3}
\end{split}\end{align}where $\frac{d\boldsymbol{\mu}(t)}{dt}$ and $\frac{d\exp(\boldsymbol{\sigma}(t))}{dt}$ are defined in Eq.~\eqref{eq1:4} as follows:
\begin{align}
    \frac{d\boldsymbol{\mu}(t)}{dt} &= g_\mu(\boldsymbol{e}(t), t; \boldsymbol{\theta}_\mu) \frac{dX(t)}{dt},\\
    \frac{d\exp(\boldsymbol{\sigma}(t))}{dt} &= \exp(\boldsymbol{\sigma}(t))\frac{d\boldsymbol{\sigma}(t)}{dt} \\
    &= \exp(\boldsymbol{\sigma}(t))\Big(g_\sigma(\boldsymbol{e}(t), t; \boldsymbol{\theta}_\sigma) \frac{dX(t)}{dt}\Big).
\end{align}

\paragraph{\textbf{Network architecture:}}The architecture of the NCDE function $k$ in the decoder is as follows:
\begin{equation}
   \begin{aligned}
        k(\boldsymbol{d}(t); \boldsymbol{\theta}_k) &= \psi(\texttt{FC}(\mathbf{D}_L)) \\
        \vdots \\
        \mathbf{D}_1 &= \omega(\texttt{FC}({\mathbf{D}_0})),\\
        \mathbf{D}_0 &= \omega(\texttt{FC}({\mathbf{d}(t)})),
    \end{aligned} 
\end{equation}where we use $dim(\mathbf{h})$ to denote the hidden size before the final layer and $dim(\mathbf{l})$ to denote the size of the final layer. Therefore, $\mathbf{D}_i$ has a size of $dim(\mathbf{h})$ for all $i$ and the output size of $k$ is $dim(\mathbf{l})$.

\begin{figure*}[t]
    \centering
    \includegraphics[width=0.8\textwidth]{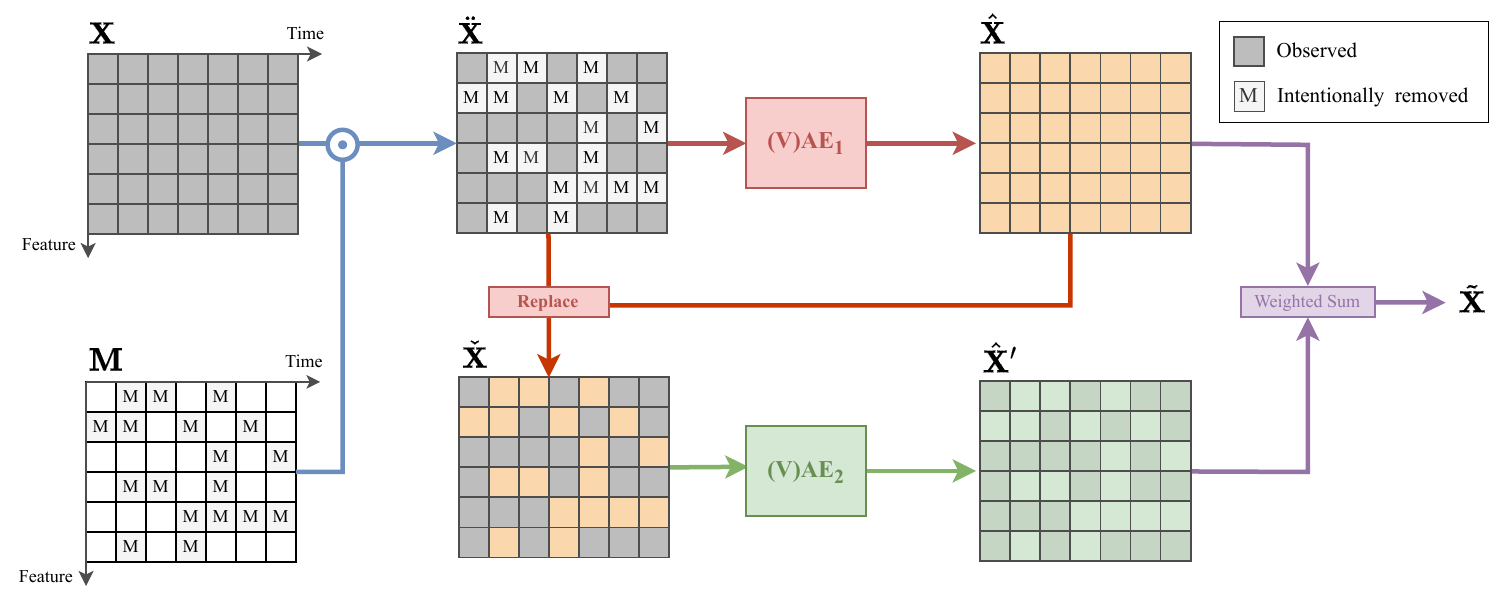}
    \caption{Overall workflow of our proposed dual-layer imputation method. We intentionally remove some elements in $\ddot{\mathbf{X}}$ for the training effect in challenging environments.}
    \label{fig:mit}
\end{figure*}

\subsection{Output Layer}
In order to infer an observation $\hat{\mathbf{x}}_i$ at time $t_i$, we use the following output layer:
\begin{align}
\hat{\mathbf{x}}_i = \texttt{FC}_2(\texttt{ELU}(\texttt{FC}_1(\boldsymbol{d}(t_i)))),
\end{align}where \texttt{FC} means an fully-connected layer, and \texttt{ELU} means an exponential linear unit. Taking the elements of $\hat{\mathbf{x}}_i$ whose original values are \textit{nan} in $\mathbf{x}_i$, we can accomplish the time series imputation task.


\subsection{Augmented ODE for Encoder and Decoder}
In order to implement our model, we use the following augmented ordinary differential equation (ODE):

\begin{align} \label{eq3}
\frac{d}{dt}{\begin{bmatrix}
  \boldsymbol{\mu}(t) \\
  \boldsymbol{\sigma}(t) \\
  \boldsymbol{d}(t) \\
  \end{bmatrix}\!} = {\begin{bmatrix}
  g_\mu(\boldsymbol{e}(t), t; \boldsymbol{\theta}_\mu) \frac{dX(t)}{dt} \\
  g_\sigma(\boldsymbol{e}(t), t; \boldsymbol{\theta}_\sigma) \frac{dX(t)}{dt} \\
  k(\boldsymbol{d}(t); \boldsymbol{\theta}_k) \Big(\frac{d\boldsymbol{\mu}(t)}{dt} + \boldsymbol{\epsilon}_t\odot \frac{d\exp(\boldsymbol{\sigma}(t))}{dt}\Big)\\
  \end{bmatrix}.\!}
\end{align} and 
\begin{align*}
{\begin{bmatrix}
  \boldsymbol{\mu}(0) \\
  \boldsymbol{\sigma}(0) \\
  \boldsymbol{d}(0) \\
  \end{bmatrix}\!} = {\begin{bmatrix}
  \texttt{FC}_{\mu}(X(0)) \\
  \texttt{FC}_{\sigma}(X(0)) \\
  \texttt{FC}_{\boldsymbol{d}}(X(0)) \\
  \end{bmatrix}.\!}
\end{align*}

Throughout Eq.~\eqref{eq3}, we can integrate our proposed continuous-time encoder and decode into a single ODE state, which means that by solving the ODE, the entire forward pass of our continuous-time variational autoencoder can be calculated simultaneously.



\paragraph{\textbf{Why continuous hidden path?:}}
We note that the hidden representation $H(t)$ in Eq.~\eqref{eq1:5} is continuously defined over time, which is different from existing methods where only a single hidden representation is created after reading the entire time series (cf. Fig.~\ref{fig:latentODE} vs.~\ref{fig:CTA}). The benefits of our proposed continuous hidden path are two folds.

Firstly, our proposed method is suitable for time series imputation. For instance, suppose that we want to infer $(\hat{\mathbf{x}}_j, t_j)$ for time series imputation. $H(t_j)$ contains the information of the input time series up to time $t$ and its near future --- note that additional information around time point $t_j$ is used when creating $X(t_j)$ with an interpolation method (cf. Fig~\ref{fig:nerualcde2}). Thus, $H(t_j)$ contains enough information to infer $\hat{\mathbf{x}}_j$ via the decoder and the output layer.

Secondly, our proposed method provides one-way lightweight processing. Only by solving the augmented ODE in Eq.~\eqref{eq3} from an initial time $0$ to a terminal time $T$ sequentially and incrementally, we can impute all missing elements with the output layer.

\paragraph{\textbf{Vanilla autoencoder:}}
Our framework can be converted to the vanilla autoencoder in a na\"ive way only by i) setting $H(t) = \boldsymbol{\mu}(t)$ after removing $\boldsymbol{\sigma}(t)$ and ii) using the usual reconstruction loss (without the ELBO~\cite{kingma2013auto} loss). Since we discard $\boldsymbol{\sigma}(t)$ in this vanilla setting, its inference time and space complexities are reduced in comparison with those of the variational autoencoder setting.

\paragraph{\textbf{How to infer:}} For inference, we use only $\boldsymbol{\mu}(t)$, i.e., $H(t) = \boldsymbol{\mu}(t)$ is used for the variational autoencoder setting. In other words, we use the mean hidden representation only. By considering $\boldsymbol{\sigma}(t)$, we can further extract the confidence interval, but our main interest is how to impute incomplete time series. For the vanilla setting, we remove $\boldsymbol{\sigma}(t)$ so it clear that $H(t) = \boldsymbol{\mu}(t)$ for inference.



\subsection{Dual Autoencoder Architecture}
We have described how a single (variational) autoencoder can be defined so far. For time series imputation, however, two-layer architectures are popular~\cite{saits}. We also propose the following dual-autoencoder approach and its training method (cf. Fig.~\ref{fig:mit}):
\begin{enumerate}
\item ({\color{blue}Blue Path} of Fig.~\ref{fig:mit}) Given a training time series sample $\mathbf{X}$, i.e., a matrix representation of $\{(\mathbf{x}_i, t_i)\}_{i=0}^N$, we intentionally remove some more elements from $\mathbf{X}$ in order to create more challenging training environments. The intentionally removed elements are marked as `M' in Fig.~\ref{fig:mit}, and we use $\mathbf{M}$ to denote the masking matrix, e.g., 1 in $\mathbf{M}$ means `intentionally removed by us.' We use $N_M$ to denote the number of these intentionally removed elements, which is a hyperparameter for our training method.
\item ({\color{red}Red Path} of Fig.~\ref{fig:mit}) We first take $\hat{\mathbf{X}}$, i.e., $\{(\hat{\mathbf{x}}_{t_i}, t_i)\}_{i=0}^N$, from the initial proposed (variational) autoencoder marked as `$(V)AE_1$' in Fig.~\ref{fig:mit}. We then create the initial imputation outcome $\check{\mathbf{X}}$ by replacing the missing \textit{nan} elements of $\ddot{\mathbf{x}}_{t_i}$ with the inferred elements of $\hat{\mathbf{x}}_{t_i}$ for all $i$.
\item ({\color{lime}Green Path} of Fig.~\ref{fig:mit}) We then feed $\check{\mathbf{X}}$ to the next proposed (variational) autoencoder. Let $\hat{\mathbf{X}}'$, i.e., $\{(\hat{\mathbf{x}}'_i, t_i)\}_{i=0}^N$, be the output from the second autoendoer via the residual connection with $\check{\mathbf{X}}$.
\item ({\color{purple}Purple Path} of Fig.~\ref{fig:mit}) We then let $\tilde{\mathbf{X}}$, i.e., $\{(\tilde{\mathbf{x}}_j,t_j)\}_{i=0}^N$, be our final imputation outcome, which is calculated as follows --- in other words, the first and second imputation outcomes are connected through the \emph{learnable} weighted sum:
\begin{align}
    \tilde{\mathbf{x}}_j = \boldsymbol\alpha\odot \hat{\mathbf{x}}_j + (\mathbf{1}-\boldsymbol\alpha)\odot\hat{\mathbf{x}}'_j,
\end{align}where $\boldsymbol\alpha = \phi(\texttt{FC}_3(\boldsymbol{d}'(t_j)))$, $\phi$ is Sigmoid, and $\boldsymbol{d}'(t_j)$ means the hidden representation of the decoder of the 2nd autoencoder at time $t_j$. $\odot$ means the element-wise multiplication.
\end{enumerate}

\begin{algorithm}[t]
\SetAlgoLined
\caption{How to train CTA}\label{alg:train}
\KwIn{Training data $D_{train}$, Validating data $D_{val}$, Maximum iteration numbers $max\_iter$}
Initialize the model parameter $\Theta$ which includes $\boldsymbol{\theta}_\mu, \boldsymbol{\theta}_\sigma, \boldsymbol{\theta}_k$ and additional parameters for other layers\;

$i \gets 0$;

\While {$i < max\_iter$}{
    Create a mini-batch $B := \{(\mathbf{X}_b, \ddot{\mathbf{X}}_b)\}_{b=1}^{N_B}$\;\label{alg:train2}
    Train $\Theta$ using the loss $L$ with $B$\;\label{alg:train3}
    Validate and update the best parameter $\Theta$ with $D_{val}$\;
    
    $i \gets i + 1$;
}
\Return $\Theta$;
\end{algorithm}
\vspace{-0.5cm}
\paragraph{\textbf{Training method:}} In order to train the dual autoencoders, we use the following loss function and the method in Alg.~\ref{alg:train}:
\begin{align}\begin{split}\label{eq:loss}
L := &\overset{\textrm{Reconstruction}}{\overbrace{||(\mathbf{X} - \tilde{\mathbf{X}})||_F + ||(\mathbf{X} - \hat{\mathbf{X}})||_F + ||(\mathbf{X} - \hat{\mathbf{X}}')||_F + ||\mathbf{M} \odot (\mathbf{X} - \tilde{\mathbf{X}})||_F}}\\
&+ \overset{\textrm{KL Divergence}}{\overbrace{\int_0^T KLD_1(t) + KLD_2(t) dt}},
\end{split}\end{align}where we use $KLD_1(t)$ and $KLD_2(t)$ for brevity to denote the usual KL divergence~\cite{csiszar1975divergence} terms of the first and the second variational autoencoders at time $t$, respectively --- note that those KL Divergence terms can be omitted for the vanilla setting of our proposed method. In particular, those KLD terms can be defined for every time point $t$ since our CTA is a method to create an infinite number of variational autoencoders in $[0,T]$ (cf. Fig.~\ref{fig:infinite_autoencoder}) and therefore, we need to integrate them over time. We use existing ODE solvers for this purpose as we do it for NCDEs (see Appendix~\ref{a:ode} for details). We also note that our loss is a continuous generalization of the ELBO loss since we have an infinite number of variational autoencoders in the time domain $[0,T]$ and therefore, the KLD loss term is defined for every time point $t \in [0,T]$.

In Alg.~\ref{alg:train}, we first initialize all the model parameters, denoted $\Theta$. In Line~\ref{alg:train2}, we create a mini-batch of size $N_B$. Each pair of $(\mathbf{X}_b, \ddot{\mathbf{X}}_b)$ means an incomplete time series sample $\ddot{\mathbf{X}}_b$ and its ground-truth sample $\mathbf{X}_b$. In Line~\ref{alg:train3}, we train $\Theta$ following the method described for Fig.~\ref{fig:mit}. As described, our training process intentionally removes $N_M$ more elements from each $\ddot{\mathbf{X}}_b$ to increase the effect of the supervised training. Our CTA produces two intermediate and one final inference outcomes, i.e., $\tilde{\mathbf{X}}_b, \hat{\mathbf{X}}_b$, and $\hat{\mathbf{X}}'_b$, for each training sample $\ddot{\mathbf{X}}_b$, with which the training with the loss $L$ is conducted.

\paragraph{\textbf{Role of each layer:}} In the proposed dual autoencoder architecture, the first (variational) autoencoder infers the initial imputed time series where for some challenging imputation points, its quality may not be satisfactory. The second (variational) autoencoder then tries to complement for the challenging cases via the learnable weighted sum, i.e., the learnable residual connection. In ablation study, we analyze the benefits of the dual-layer architecture.

\vspace{-0.1cm}
\paragraph{\textbf{How to Solve $\int_0^T KLD(t) dt$?:}}\label{a:ode}
Our loss function in Eq.~\eqref{eq:loss} requires an integral problem to calculate the KLD terms along the time in $[0,T]$. For simplicity but without loss of generality, we assume only one variational autoencoder so we need to solve $\int_0^T KLD(t) dt$. For this, we define and solve the following augmented ODE to calculate all the hidden states and the KLD loss at the same time. Therefore, $\xi(T)$ corresponds to the Riemann integral of $\int_0^T KLD(t) dt$ and contains the final KLD loss value.
\begin{align*} 
\frac{d}{dt}{\begin{bmatrix}
  \boldsymbol{\mu}(t) \\
  \boldsymbol{\sigma}(t) \\
  \boldsymbol{d}(t) \\
  \xi(t)
  \end{bmatrix}\!} = {\begin{bmatrix}
  g_\mu(\boldsymbol{e}(t), t; \boldsymbol{\theta}_\mu) \frac{dX(t)}{dt} \\
  g_\sigma(\boldsymbol{e}(t), t; \boldsymbol{\theta}_\sigma) \frac{dX(t)}{dt} \\
  k(\boldsymbol{d}(t); \boldsymbol{\theta}_k) \Big(\frac{d\boldsymbol{\mu}(t)}{dt} + \boldsymbol{\epsilon}_t\odot \frac{d\exp(\boldsymbol{\sigma}(t))}{dt}\Big)\\
  KLD(t)
  \end{bmatrix}.\!}
\end{align*}

\vspace{-0.2cm}

\begin{table*}[t]
\centering
\setlength{\tabcolsep}{1.5pt}
\caption{Performance on \texttt{AirQuality} and \texttt{Stocks}. The best results are in boldface and the second best results are underlined. Lower errors are preferred. SA-based$(^\ast)$ indicates self-attention-based model.}
\label{table:main_air_stock}
\resizebox{0.95\textwidth}{!}{%
    \begin{tabular}{c|c|cc|cc|cc|cc|cc|cc}
    \specialrule{1pt}{1pt}{1pt}
    \multicolumn{2}{c|}{Dataset} & \multicolumn{6}{c|}{\texttt{AirQuality}}&\multicolumn{6}{c}{\texttt{Stocks}} \\
    \specialrule{0.5pt}{0.5pt}{0.5pt}
    \multicolumn{2}{c|}{\multirow{2}{*}{\footnotesize\diagbox[width=15em]{Method}{$r_{missing}$}}}   & \multicolumn{2}{c|}{30\%}  & \multicolumn{2}{c|}{50\%}      & \multicolumn{2}{c|}{70\%}  & \multicolumn{2}{c|}{30\%}  & \multicolumn{2}{c|}{50\%}      & \multicolumn{2}{c}{70\%} \\
    \cline{3-14}
    \multicolumn{2}{c|}{} & \footnotesize{MAE} & \footnotesize{RMSE}& \footnotesize{MAE} & \footnotesize{RMSE} & \footnotesize{MAE} & \footnotesize{RMSE} & \footnotesize{MAE} & \footnotesize{RMSE} & \footnotesize{MAE} & \footnotesize{RMSE} & \footnotesize{MAE} & \footnotesize{RMSE}      \\
    \specialrule{1pt}{1pt}{1pt}
    \multirow{2}{*}{Statistical}        & KNN        &0.328{\std{±0.001}} &0.621{\std{±0.018}} &0.352{\std{±0.000}} &0.673{\std{±0.014}} &0.503{\std{±0.000}} &0.859{\std{±0.005}} &0.614{\std{±0.061}} &2.081{\std{±0.218}} &0.612{\std{±0.046}} &1.963{\std{±0.175}} &0.754{\std{±0.013}} &2.066{\std{±0.116}}  \\
                                        & MICE        &2.377{\std{±0.010}} &6.073{\std{±0.030}} &0.494{\std{±0.004}} &1.110{\std{±0.016}} &0.335{\std{±0.000}} &0.654{\std{±0.007}} &0.755{\std{±0.080}} &2.254{\std{±0.224}} &0.752{\std{±0.035}} &2.091{\std{±0.138}} &0.990{\std{±0.015}} &2.243{\std{±0.110}}  \\
    \specialrule{0.5pt}{0.5pt}{0.5pt}
    \multirow{4}{*}{RNN-based}          & GRU-D        &0.459{\std{±0.016}} &0.740{\std{±0.020}} &0.466{\std{±0.014}} &0.761{\std{±0.017}} &0.482{\std{±0.013}} &0.780{\std{±0.016}} &0.868{\std{±0.009}} &2.222{\std{±0.003}} &0.735{\std{±0.005}} &1.743{\std{±0.005}} &0.936{\std{±0.012}} &2.015{\std{±0.005}}  \\
                                        & ODE-RNN        &0.370{\std{±0.000}} &0.611{\std{±0.000}} &0.371{\std{±0.000}} &0.632{\std{±0.000}} &0.383{\std{±0.000}} &0.650{\std{±0.000}} &0.740{\std{±0.000}} &1.919{\std{±0.000}} &0.718{\std{±0.000}} &1.720{\std{±0.000}} &0.857{\std{±0.000}} &1.873{\std{±0.000}}  \\
                                        & MRNN        &0.309{\std{±0.001}} &0.566{\std{±0.019}} &0.325{\std{±0.000}} &0.585{\std{±0.014}} &0.352{\std{±0.000}} &0.614{\std{±0.006}} &0.789{\std{±0.062}} &2.170{\std{±0.217}} &0.745{\std{±0.041}} &2.101{\std{±0.168}} &0.766{\std{±0.020}} &2.086{\std{±0.118}}  \\
                                        & BRITS        &0.212{\std{±0.001}} &0.470{\std{±0.015}} &0.231{\std{±0.000}} &0.494{\std{±0.013}} &0.253{\std{±0.000}} &0.527{\std{±0.007}} &0.413{\std{±0.048}} &1.635{\std{±0.210}} &0.462{\std{±0.042}} &1.651{\std{±0.182}} &0.600{\std{±0.023}} &1.966{\std{±0.126}}  \\
    \specialrule{0.5pt}{0.5pt}{0.5pt}
    \multirow{6}{*}{VAE-based}          & RNN-VAE        &0.704{\std{±0.000}} &1.009{\std{±0.001}} &0.704{\std{±0.001}} &1.019{\std{±0.001}} &0.702{\std{±0.000}} &1.016{\std{±0.000}} &1.576{\std{±0.011}} &2.529{\std{±0.032}} &1.427{\std{±0.008}} &2.173{\std{±0.031}} &1.509{\std{±0.003}} &2.274{\std{±0.005}}  \\
                                        & LatentODE        &0.466{\std{±0.000}} &0.723{\std{±0.000}} &0.453{\std{±0.001}} &0.732{\std{±0.001}} &0.464{\std{±0.000}} &0.739{\std{±0.000}} &0.562{\std{±0.003}} &1.802{\std{±0.005}} &0.461{\std{±0.002}} &1.515{\std{±0.006}} &0.534{\std{±0.001}} &1.636{\std{±0.002}}  \\
                                        & VRNN        &0.450{\std{±0.001}} &0.744{\std{±0.016}} &0.510{\std{±0.001}} &0.803{\std{±0.011}} &0.584{\std{±0.000}} &0.880{\std{±0.004}} &1.047{\std{±0.061}} &2.284{\std{±0.214}} &1.152{\std{±0.044}} &2.197{\std{±0.166}} &1.404{\std{±0.021}} &2.396{\std{±0.109}}  \\
                                        & SRNN        &0.366{\std{±0.001}} &0.632{\std{±0.019}} &0.460{\std{±0.001}} &0.734{\std{±0.012}} &0.560{\std{±0.000}} &0.846{\std{±0.005}} &0.782{\std{±0.055}} &1.984{\std{±0.214}} &1.063{\std{±0.043}} &2.102{\std{±0.169}} &1.364{\std{±0.022}} &2.367{\std{±0.110}}  \\
                                        & STORN        &0.450{\std{±0.001}} &0.743{\std{±0.016}} &0.509{\std{±0.001}} &0.799{\std{±0.011}} &0.583{\std{±0.000}} &0.879{\std{±0.005}} &1.037{\std{±0.060}} &2.237{\std{±0.212}} &1.133{\std{±0.041}} &2.161{\std{±0.166}} &1.386{\std{±0.022}} &2.385{\std{±0.111}}  \\
                                        & GP-VAE        &0.287{\std{±0.001}} &0.517{\std{±0.015}} &0.303{\std{±0.001}} &0.547{\std{±0.013}} &0.307{\std{±0.001}} &0.556{\std{±0.005}} &0.494{\std{±0.049}} &1.574{\std{±0.211}} &0.531{\std{±0.044}} &1.503{\std{±0.194}} &0.560{\std{±0.029}} &1.604{\std{±0.216}}  \\
    \specialrule{0.5pt}{0.5pt}{0.5pt}
    \multirow{3}{*}{GAN-based}          & GRUI-GAN        &0.851{\std{±0.003}} &1.165{\std{±0.010}} &0.843{\std{±0.002}} &1.154{\std{±0.008}} &0.820{\std{±0.002}} &1.128{\std{±0.005}} &0.766{\std{±0.060}} &2.035{\std{±0.209}} &1.307{\std{±0.066}} &2.342{\std{±0.179}} &1.845{\std{±0.020}} &2.620{\std{±0.100}}  \\
                                        & E2GAN        &0.742{\std{±0.001}} &1.056{\std{±0.012}} &0.746{\std{±0.001}} &1.060{\std{±0.009}} &0.747{\std{±0.001}} &1.074{\std{±0.004}} &1.556{\std{±0.050}} &2.641{\std{±0.176}} &1.533{\std{±0.040}} &2.430{\std{±0.150}} &1.511{\std{±0.023}} &2.468{\std{±0.106}}  \\
                                        & GAIN        &0.440{\std{±0.001}} &0.680{\std{±0.014}} &0.555{\std{±0.000}} &0.808{\std{±0.011}} &0.657{\std{±0.000}} &0.938{\std{±0.005}} &1.127{\std{±0.057}} &2.267{\std{±0.205}} &1.187{\std{±0.048}} &2.147{\std{±0.166}} &1.421{\std{±0.013}} &2.442{\std{±0.094}}  \\
    \specialrule{0.5pt}{0.5pt}{0.5pt}
    \multirow{4}{*}{SA-based($^\ast$)}       & mTAN        &0.257{\std{±0.000}} &0.497{\std{±0.002}} &0.273{\std{±0.000}} &0.518{\std{±0.007}} &0.289{\std{±0.000}} &0.556{\std{±0.002}} &0.390{\std{±0.005}} &\cellcolor{gray!20} \textbf{1.045{\std{±0.020}}}
    &\underline{0.345{\std{±0.006}}} &\underline{1.082{\std{±0.013}}} &0.497{\std{±0.016}} &\underline{1.576{\std{±0.012}}}  \\
                                        & HetVAE        &0.243{\std{±0.000}} &0.505{\std{±0.021}} &0.251{\std{±0.000}} &0.518{\std{±0.016}} &0.281{\std{±0.000}} &0.557{\std{±0.006}} &\underline{0.319{\std{±0.046}}} &1.345{\std{±0.229}} &0.384{\std{±0.037}} &1.500{\std{±0.176}} &\underline{0.421{\std{±0.024}}} &1.612{\std{±0.146}}  \\
                                        & Transformer        &0.222{\std{±0.000}} &0.475{\std{±0.017}} &0.235{\std{±0.001}} &0.494{\std{±0.013}} &0.254{\std{±0.000}} &0.523{\std{±0.008}} &0.388{\std{±0.047}} &1.476{\std{±0.230}} &0.378{\std{±0.036}} &1.402{\std{±0.186}} &0.489{\std{±0.024}} &1.764{\std{±0.131}}  \\
                                        & SAITS        &\underline{0.201{\std{±0.000}}} &\underline{0.449{\std{±0.016}}} &\underline{0.230{\std{±0.001}}} &\underline{0.493{\std{±0.012}}} &\underline{0.247{\std{±0.000}}} &\underline{0.513{\std{±0.006}}} &0.374{\std{±0.048}} &1.461{\std{±0.233}} &0.371{\std{±0.036}} &1.369{\std{±0.195}} &0.474{\std{±0.026}} &1.733{\std{±0.129}}  \\
    \specialrule{1pt}{1pt}{1pt}
    \multirow{2}{*}{CTA (ours)}         & VAE-AE        & \cellcolor{gray!20} \textbf{0.186{\std{±0.000}}} & \cellcolor{gray!20} \textbf{0.424{\std{±0.023}}} & \cellcolor{gray!20} \textbf{0.202{\std{±0.000}}} & \cellcolor{gray!20} \textbf{0.447{\std{±0.019}}} &0.237{\std{±0.000}} &0.513{\std{±0.007}} & \cellcolor{gray!20} \textbf{0.289{\std{±0.034}}} & \underline{1.138{\std{±0.186}}} & \cellcolor{gray!20} \textbf{0.281{\std{±0.035}}} & \cellcolor{gray!20} \textbf{1.075\std{±0.191}} & \cellcolor{gray!20} \textbf{0.335{\std{±0.008}}} & \cellcolor{gray!20} \textbf{1.283{\std{±0.107}}}  \\
                                        & AE-AE        &0.200{\std{±0.001}} &0.438{\std{±0.020}} &0.217{\std{±0.000}} &0.481{\std{±0.018}} &\cellcolor{gray!20} \textbf{0.235{\std{±0.000}}} &\cellcolor{gray!20} \textbf{0.508{\std{±0.008}}} &0.300{\std{±0.035}} &1.204{\std{±0.160}} &0.315{\std{±0.020}} &1.202{\std{±0.108}} &0.364{\std{±0.009}} &1.365{\std{±0.074}}  \\
    \specialrule{1pt}{1pt}{1pt}
    \end{tabular}%
}
\end{table*}

\paragraph{\textbf{Original Missing Elements of \ $\mathbf{X}$}:}\label{a:missing}
For ease of our discussion, we assumed that for $\mathbf{X}$, all ground-truth elements are known in the main body of this paper. In our experiments, however, some ground-truth elements are unknown in their originally released dataset. In this situation, we cannot use those elements for training and testing. Modifying our descriptions in the main paper to consider those missing ground-truth elements is straightforward. For instance, $\boldsymbol\alpha$ is redefined to $\boldsymbol\alpha = \phi(\texttt{FC}_3(\boldsymbol{d}'(t_j), \mathbf{O}))$ and the loss function can be rewritten as follows:
\begin{align*}\begin{split}
L := &\overset{\textrm{Reconstruction}}{\overbrace{||\mathbf{O} \odot (\mathbf{X} - \tilde{\mathbf{X}})||_F + ||\mathbf{O} \odot (\mathbf{X} - \hat{\mathbf{X}})||_F + ||\mathbf{O} \odot (\mathbf{X} - \hat{\mathbf{X}}')||_F}}\\
&+ \overset{\textrm{Reconstruction}}{\overbrace{||\mathbf{M} \odot (\mathbf{X} - \tilde{\mathbf{X}})||_F}} + \overset{\textrm{KL Divergence}}{\overbrace{\int_0^T KLD_1(t) + KLD_2(t) dt}},
\end{split}\end{align*}where $\mathbf{O}$ means a masking matrix to denote those elements whose ground-truth values are known in its original dataset.

\begin{table*}[t]
\centering
\setlength{\tabcolsep}{1.5pt}
\caption{Performance on \texttt{Electricity} and \texttt{Energy}. The best results are in boldface and the second best results are underlined. Lower errors are preferred. SA-based$(^\ast)$ indicates self-attention-based model.}
\label{table:main_elec_energy}
\resizebox{0.95\textwidth}{!}{%
    \begin{tabular}{c|c|cc|cc|cc|cc|cc|cc}
    \specialrule{1pt}{1pt}{1pt}
    \multicolumn{2}{c|}{Dataset} & \multicolumn{6}{c|}{\texttt{Electricity}}&\multicolumn{6}{c}{\texttt{Energy}} \\
    \specialrule{0.5pt}{0.5pt}{0.5pt}
    \multicolumn{2}{c|}{\multirow{2}{*}{\footnotesize\diagbox[width=15em]{Method}{$r_{missing}$}}}   & \multicolumn{2}{c|}{30\%}  & \multicolumn{2}{c|}{50\%}      & \multicolumn{2}{c|}{70\%}  & \multicolumn{2}{c|}{30\%}  & \multicolumn{2}{c|}{50\%}      & \multicolumn{2}{c}{70\%} \\
    \cline{3-14}
    \multicolumn{2}{c|}{} & \footnotesize{MAE} & \footnotesize{RMSE}& \footnotesize{MAE} & \footnotesize{RMSE} & \footnotesize{MAE} & \footnotesize{RMSE} & \footnotesize{MAE} & \footnotesize{RMSE} & \footnotesize{MAE} & \footnotesize{RMSE} & \footnotesize{MAE} & \footnotesize{RMSE}      \\
    \specialrule{1pt}{1pt}{1pt}
    \multirow{2}{*}{Statistical}        & KNN        &1.369{\std{±0.000}} &2.047{\std{±0.002}} &1.356{\std{±0.000}} &2.060{\std{±0.001}} &1.421{\std{±0.000}} &2.137{\std{±0.001}} &0.583{\std{±0.006}} &0.787{\std{±0.009}} &0.819{\std{±0.003}} &1.071{\std{±0.005}} &1.109{\std{±0.005}} &1.404{\std{±0.005}}  \\
                                        & MICE        &\underline{0.867{\std{±0.000}}} &1.398{\std{±0.001}} &\underline{0.906{\std{±0.000}}} &1.559{\std{±0.001}} &\underline{0.972{\std{±0.000}}} &1.662{\std{±0.001}} &0.513{\std{±0.005}} &0.761{\std{±0.009}} &0.543{\std{±0.003}} &0.743{\std{±0.004}} &0.829{\std{±0.003}} &1.075{\std{±0.005}}  \\
    \specialrule{0.5pt}{0.5pt}{0.5pt}
    \multirow{4}{*}{RNN-based}          & GRU-D        &1.564{\std{±0.003}} &2.180{\std{±0.004}} &1.586{\std{±0.005}} &2.200{\std{±0.008}} &1.612{\std{±0.005}} &2.211{\std{±0.007}} &0.537{\std{±0.009}} &0.713{\std{±0.010}} &0.544{\std{±0.010}} &0.727{\std{±0.012}} &0.562{\std{±0.009}} &0.755{\std{±0.012}}  \\
                                        & ODE-RNN        &1.539{\std{±0.000}} &2.169{\std{±0.000}} &1.529{\std{±0.000}} &2.150{\std{±0.000}} &1.519{\std{±0.000}} &2.137{\std{±0.000}} &0.517{\std{±0.000}} &0.698{\std{±0.000}} &0.541{\std{±0.000}} &0.728{\std{±0.000}} &0.574{\std{±0.000}} &0.767{\std{±0.000}}  \\
                                        & MRNN        &1.272{\std{±0.000}} &1.900{\std{±0.002}} &1.297{\std{±0.000}} &1.925{\std{±0.001}} &1.325{\std{±0.000}} &1.944{\std{±0.001}} &0.555{\std{±0.003}} &0.736{\std{±0.006}} &0.593{\std{±0.002}} &0.781{\std{±0.005}} &0.654{\std{±0.002}} &0.842{\std{±0.003}}  \\
                                        & BRITS        &0.915{\std{±0.000}} &1.510{\std{±0.002}} &0.980{\std{±0.000}} &1.602{\std{±0.001}} &1.110{\std{±0.000}} &1.737{\std{±0.001}} &\underline{0.172{\std{±0.004}}} &\underline{0.324{\std{±0.011}}} &\underline{0.278{\std{±0.004}}} &\underline{0.438{\std{±0.009}}} &0.411{\std{±0.001}} &\underline{0.589{\std{±0.005}}}  \\
    \specialrule{0.5pt}{0.5pt}{0.5pt}
    \multirow{6}{*}{VAE-based}          & RNN-VAE        &1.864{\std{±0.000}} &2.447{\std{±0.000}} &1.849{\std{±0.000}} &2.413{\std{±0.000}} &1.846{\std{±0.000}} &2.417{\std{±0.000}} &0.965{\std{±0.001}} &1.211{\std{±0.000}} &0.973{\std{±0.000}} &1.220{\std{±0.001}} &0.968{\std{±0.000}} &1.213{\std{±0.000}}  \\
                                        & LatentODE        &1.814{\std{±0.001}} &2.450{\std{±0.001}} &1.732{\std{±0.001}} &2.352{\std{±0.001}} &1.777{\std{±0.001}} &2.404{\std{±0.001}} &0.712{\std{±0.002}} &0.935{\std{±0.003}} &0.697{\std{±0.008}} &0.912{\std{±0.009}} &0.698{\std{±0.008}} &0.916{\std{±0.009}}  \\
                                        & VRNN        &1.565{\std{±0.000}} &2.159{\std{±0.002}} &1.635{\std{±0.000}} &2.224{\std{±0.001}} &1.705{\std{±0.000}} &2.281{\std{±0.001}} &0.896{\std{±0.004}} &1.141{\std{±0.004}} &0.920{\std{±0.001}} &1.161{\std{±0.003}} &0.933{\std{±0.001}} &1.172{\std{±0.002}}  \\
                                        & SRNN        &1.534{\std{±0.000}} &2.063{\std{±0.001}} &1.636{\std{±0.000}} &2.185{\std{±0.001}} &1.700{\std{±0.000}} &2.256{\std{±0.001}} &0.537{\std{±0.004}} &0.717{\std{±0.007}} &0.654{\std{±0.002}} &0.846{\std{±0.005}} &0.783{\std{±0.001}} &0.995{\std{±0.003}}  \\
                                        & STORN        &1.532{\std{±0.000}} &2.098{\std{±0.001}} &1.622{\std{±0.000}} &2.188{\std{±0.001}} &1.702{\std{±0.000}} &2.265{\std{±0.001}} &0.869{\std{±0.005}} &1.100{\std{±0.006}} &0.895{\std{±0.002}} &1.132{\std{±0.003}} &0.915{\std{±0.001}} &1.152{\std{±0.002}}  \\
                                        & GP-VAE        &1.006{\std{±0.000}} &1.633{\std{±0.001}} &1.058{\std{±0.000}} &1.704{\std{±0.001}} &1.119{\std{±0.000}} &1.764{\std{±0.001}} &0.407{\std{±0.001}} &0.551{\std{±0.004}} &0.496{\std{±0.003}} &0.654{\std{±0.006}} &0.504{\std{±0.001}} &0.668{\std{±0.002}}  \\
    \specialrule{0.5pt}{0.5pt}{0.5pt}
    \multirow{3}{*}{GAN-based}          & GRUI-GAN        &1.919{\std{±0.001}} &2.510{\std{±0.002}} &1.905{\std{±0.001}} &2.508{\std{±0.002}} &1.894{\std{±0.001}} &2.487{\std{±0.001}} &0.958{\std{±0.004}} &1.171{\std{±0.005}} &1.057{\std{±0.007}} &1.329{\std{±0.006}} &0.999{\std{±0.005}} &1.262{\std{±0.007}}  \\
                                        & E2GAN        &1.883{\std{±0.000}} &2.459{\std{±0.001}} &1.874{\std{±0.000}} &2.442{\std{±0.001}} &1.880{\std{±0.000}} &2.457{\std{±0.001}} &0.856{\std{±0.005}} &1.088{\std{±0.003}} &0.843{\std{±0.002}} &1.066{\std{±0.002}} &0.864{\std{±0.002}} &1.103{\std{±0.003}}  \\
                                        & GAIN        &1.304{\std{±0.000}} &1.832{\std{±0.001}} &1.546{\std{±0.000}} &2.109{\std{±0.001}} &1.733{\std{±0.000}} &2.288{\std{±0.001}} &0.647{\std{±0.004}} &0.842{\std{±0.006}} &0.811{\std{±0.001}} &1.033{\std{±0.004}} &0.917{\std{±0.002}} &1.148{\std{±0.004}}  \\
    \specialrule{0.5pt}{0.5pt}{0.5pt}
    \multirow{4}{*}{SA-based($^\ast$)}       & mTAN        &1.326{\std{±0.000}} &1.840{\std{±0.000}} &1.378{\std{±0.000}} &1.887{\std{±0.000}} &1.386{\std{±0.000}} &1.962{\std{±0.000}} &0.394{\std{±0.001}} &0.553{\std{±0.001}} &0.389{\std{±0.001}} &0.549{\std{±0.001}} &0.422{\std{±0.000}} &0.595{\std{±0.000}}  \\
                                        & HetVAE        &1.168{\std{±0.001}} &1.973{\std{±0.002}} &1.238{\std{±0.000}} &1.878{\std{±0.001}} &1.176{\std{±0.000}} &1.850{\std{±0.001}} &0.316{\std{±0.004}} &0.486{\std{±0.008}} &0.335{\std{±0.002}} &0.501{\std{±0.003}} &\underline{0.409{\std{±0.002}}} &\underline{0.589{\std{±0.006}}}  \\
                                        & Transformer        &0.899{\std{±0.000}} &\underline{1.336{\std{±0.001}}} &0.946{\std{±0.000}} &\underline{1.518{\std{±0.001}}} &0.988{\std{±0.000}} &\underline{1.574{\std{±0.001}}} &0.347{\std{±0.005}} &0.515{\std{±0.014}} &0.423{\std{±0.004}} &0.601{\std{±0.008}} &0.502{\std{±0.003}} &0.705{\std{±0.006}}  \\
                                        & SAITS        &0.894{\std{±0.000}} &1.404{\std{±0.001}} &0.953{\std{±0.000}} &1.581{\std{±0.001}} &1.024{\std{±0.000}} &1.680{\std{±0.001}} &0.313{\std{±0.006}} &0.484{\std{±0.015}} &0.392{\std{±0.004}} &0.570{\std{±0.008}} &0.484{\std{±0.004}} &0.675{\std{±0.007}}  \\
    \specialrule{1pt}{1pt}{1pt}
    \multirow{2}{*}{CTA (ours)}         & VAE-AE        &0.767{\std{±0.000}} &\cellcolor{gray!20} \textbf{1.127{\std{±0.001}}} &\cellcolor{gray!20} \textbf{0.748{\std{±0.000}}} &\cellcolor{gray!20} \textbf{1.146{\std{±0.001}}} &\cellcolor{gray!20} \textbf{0.781{\std{±0.000}}} &\cellcolor{gray!20} \textbf{1.149{\std{±0.001}}} &0.205{\std{±0.004}} &0.343{\std{±0.011}} &0.227{\std{±0.001}} &0.381{\std{±0.006}} &0.287{\std{±0.002}} &0.460{\std{±0.007}}  \\
                                        & AE-AE        &\cellcolor{gray!20} \textbf{0.742{\std{±0.000}}} &1.139{\std{±0.001}} &0.772{\std{±0.000}} &1.162{\std{±0.001}} &0.810{\std{±0.000}} &1.215{\std{±0.001}} &\cellcolor{gray!20} \textbf{0.170{\std{±0.004}}} & \cellcolor{gray!20} \textbf{0.316{\std{±0.014}}} &\cellcolor{gray!20} \textbf{0.208{\std{±0.002}}} &\cellcolor{gray!20} \textbf{0.365{\std{±0.007}}} &\cellcolor{gray!20} \textbf{0.280{\std{±0.002}}} &\cellcolor{gray!20} \textbf{0.450{\std{±0.007}}}  \\
    \specialrule{1pt}{1pt}{1pt}
    \end{tabular}%
}
\end{table*}

\section{Experiments}
In this section, we describe our experimental environments followed by experimental results and analyses.

\subsection{Experimental Environments}

\subsubsection{Datasets}

To evaluate the performance of various methods, we use four real-world datasets from different domains as follows: \texttt{AirQuality}, \texttt{Stocks}, \texttt{Electricity} and \texttt{Energy} (See supplementray material for their details). 

\subsubsection{Baselines}
 We compare our method with 19 baselines, which include statistical methods, VAE-based, RNN-based, GAN-based and self-attention-based methods (see supplementray material for details).

\subsubsection{Evaluation Methods}
To evaluate our method and baselines, we utilize two metrics: MAE (Mean Absolute Error) and RMSE (Root Mean Square Error). These are commonly used in the time series imputation literature~\cite{brits,saits,shukla2021hetvae}. We report the mean and standard deviation of the error for five trials.

In order to create more challenging evaluation environments, we increase the percentage of the missing elements, denoted $r_{missing}$. We remove the element by the ratio of $r_{missing}$ from the training, i.e., the model does not learn about this missing elements, and test datasets, i.e., the imputation task's targets are those missing elements. In total, we test in three different settings, i.e., $r_{missing} \in \{ 30\%, 50\%, 70\%\}$. The above evaluation metrics are measured only for those missing elements since our task is imputation.

\subsubsection{Hyperparameters}
We report the search range of each hyperparameter in our method and all the baselines in our supplementary material. In addition, we summarize the best hyperparameter of our method for reproducibility in the supplementray material.

\subsection{Experimental Results}

Table~\ref{table:main_air_stock} summarizes the results on \texttt{AirQuality} and \texttt{Stocks}. For \texttt{AirQuality}, the performances of SAITS and BRITS are the best among the baselines for all missing rates, but CTA shows the lowest errors in all cases. 

In the case of \texttt{Stocks}, our method, the self-attention-based methods, and some of the VAE-based methods work reasonably. When $r_{missing}$ is 30\%, mTAN performs slightly better than our model in RMSE. However, in all other cases, the performance of CTA (VAE-AE) outperforms other baselines by large margins.

The results on \texttt{Electricity} and \texttt{Energy} are shown in Table~\ref{table:main_elec_energy}. For \texttt{Electricity}, MICE, which is a statistical method, shows the best result among all the baselines. However, Our CTA marks the best accuracy in general. In particular, CTA significantly outperforms others baselines when $r_{missing}$ is high.

In the case of \texttt{Energy}, BRITS shows the best performance among the baselines. However, it is shown that the error increases rapidly as $r_{missing}$ increases. When $r_{missing}$ is high, the performance of HetVAE is the best among the baselines, but our CTA (AE-AE) outperforms other baselines at all missing rates.

\begin{table}[t]
    \centering
    \vspace{0.3cm}
    \caption{The number of model parameters and the GPU memory usage for the best hyperparameter settings of CTA and SAITS with $r_{missing}$=70\%}
    \label{table:num_param}
\resizebox{0.95\columnwidth}{!}{%
    \begin{tabular}{c|c|c}
    \toprule
        {\footnotesize\diagbox[width=8em]{Dataset}{Method}}    & SAITS                 & CTA     \\
        \midrule
        \texttt{AirQuality}          & 2.35 M / 204.33 MB     & 1.52 M / 57.45 MB (AE-AE) \\
        \texttt{Stocks}              & 12.64 M / 184.89 MB     & 0.02 M / 0.42 MB (VAE-AE) \\
        \texttt{Electricity}         & 16.13 M / 1,185.44 MB     & 53.26 M / 950.85 MB (VAE-AE) \\
        \texttt{Energy}              & 11.64 M / 443.85 MB     & 0.92 M / 13.41 MB (AE-AE) \\    
    \bottomrule
    \end{tabular}%
}
\end{table}

\subsection{Empirical Complexity}
We compare the model sizes and the inference GPU memory usage of our method and SAITS, the best-performing baseline, in Table~\ref{table:num_param}. Except for the number of parameters for \texttt{Electricity}, our model has a smaller size and consumes less GPU memory than SAITS. Especially for \texttt{Stocks} and \texttt{Energy}, our model's size is 2 to 3 orders of magnitude smaller than that of SAITS, which is an outstanding result. One more interesting point is that for \texttt{Electricity}, CTA marks comparable GPU memory footprint to SAITS with more parameters, which shows the efficiency of our computation.

\label{dual autoencoders}
\begin{table}[t]

\setlength{\tabcolsep}{2.5pt}
\caption{Results with various architectures for our CTA on \texttt{AirQuality} and \texttt{Stocks} with $r_{missing}$=70\%. The best results are in boldface and the second best results are underlined.}
\label{table:ablation_dual}
\resizebox{0.95\columnwidth}{!}{%
\begin{tabular}{l|cc|cc}
\specialrule{1pt}{1pt}{1pt}
\multirow{2}{*}{{\footnotesize\diagbox[width=8.2em]{Method}{Dataset}}} & \multicolumn{2}{c|}{\texttt{AirQuality}} & \multicolumn{2}{c}{\texttt{Stocks}}      \\
\cline{2-5}
                    & \footnotesize{MAE}    & \footnotesize{RMSE}   & \footnotesize{MAE}    & \footnotesize{RMSE} \\
\specialrule{1pt}{1pt}{1pt}
        AE    & 0.240{\std{±0.000}}  & 0.513{\std{±0.008}}  &  0.655{\std{±0.182}}   & 3.369{\std{±1.122}} \\
        VAE    & 0.246{\std{±0.000}}  & 0.520{\std{±0.007}} &  0.370{\std{±0.043}}   & 1.432{\std{±0.117}} \\
        AE-AE (ours)    & \cellcolor{gray!20} \textbf{0.235{\std{±0.000}}}   & 0.508{\std{±0.008}}  &  0.364{\std{±0.009}} & \underline{1.365{\std{±0.074}}} \\
        AE-VAE    & 0.252{\std{±0.001}}   & 0.523{\std{±0.007}}  &  0.456{\std{±0.068}}   & 1.711{\std{±0.192}} \\
        VAE-AE (ours)    & 0.237{\std{±0.000}}   & 0.513{\std{±0.007}}  &  \cellcolor{gray!20} \textbf{0.335{\std{±0.008}}} & \cellcolor{gray!20} \textbf{1.283{\std{±0.107}}} \\
        VAE-VAE    & 0.238{\std{±0.000}}   & 0.514{\std{±0.008}}  &  0.406{\std{±0.041}}   & 1.584{\std{±0.171}} \\
        AE-AE-AE    & \cellcolor{gray!20} \textbf{0.235{\std{±0.000}}}   & \underline{0.504{\std{±0.005}}}  &  0.509{\std{±0.096}}   & 2.013{\std{±0.426}} \\
        AE-AE-VAE    & \underline{0.236{\std{±0.000}}}   & \cellcolor{gray!20} \textbf{0.503{\std{±0.007}}}  &  0.385{\std{±0.007}}   & 1.410{\std{±0.122}} \\
        AE-VAE-AE    & \underline{0.236{\std{±0.000}}}   & 0.507{\std{±0.008}}  &  0.370{\std{±0.010}}   & 1.391{\std{±0.097}} \\
        AE-VAE-VAE    & 0.237{\std{±0.000}}   & 0.509{\std{±0.007}}  &  \underline{0.358{\std{±0.011}}}   & 1.366{\std{±0.109}} \\
        VAE-AE-AE    & 0.237{\std{±0.000}}   & \underline{0.504{\std{±0.008}}}  &  0.389{\std{±0.011}}   & 1.404{\std{±0.105}} \\
        VAE-AE-VAE    & 0.237{\std{±0.000}}   & 0.505{\std{±0.008}}  &  0.409{\std{±0.011}}   & 1.414{\std{±0.085}} \\
        VAE-VAE-AE    & 0.238{\std{±0.000}}   & 0.511{\std{±0.007}}  &  0.465{\std{±0.049}}   & 1.743{\std{±0.195}} \\
        VAE-VAE-VAE    & 0.238{\std{±0.000}}   & 0.512{\std{±0.007}}  &  0.404{\std{±0.013}}   & 1.448{\std{±0.115}} \\
\specialrule{1pt}{1pt}{1pt}
\end{tabular}
}
\vspace{-1em}
\end{table}

\subsection{Ablation Study on Dual-Layer Architecture}
Since CTA uses a dual-layer autoencoder architecture, we conduct an ablation study by varying the number of layers. We test all the combinations from single to triple layers, and their results are shown in Table~\ref{table:ablation_dual} for \texttt{AirQuality} and \texttt{Stocks} with the 70\% missing rate. In general, VAE-AE and AE-AE show good results for our CTA. In both datasets, the single-layer ablation models, i.e., AE and VAE, produce worse outcomes than those of the dual-layer models. Among the dual-layer models, it shows better outcomes when the second layer is AE instead of VAE. For \texttt{AirQuality}, The performances of dual and triple-layer are not significantly different.
\begin{table}[!t]

\setlength{\tabcolsep}{1.5pt}
\caption{Results with the interpolation method on \texttt{AirQuality} and \texttt{Stocks} with $r_{missing}$=70\%. The best results are in boldface.}
\label{table:sensitivity_interpolation}
\resizebox{0.95\columnwidth}{!}{%
    \begin{tabular}{c|cc|cc}

    \specialrule{1pt}{1pt}{1pt}
    \multirow{2}{*}{{\footnotesize\diagbox[width=11em]{Method}{Dataset}}} & \multicolumn{2}{c|}{\texttt{AirQuality}} & \multicolumn{2}{c}{\texttt{Stocks}}      \\
    \cline{2-5}
                        & \footnotesize{MAE}    & \footnotesize{RMSE}   & \footnotesize{MAE}    & \footnotesize{RMSE} \\
    \specialrule{1pt}{1pt}{1pt}
            Natural Cubic Spline    & 0.250{\std{±0.001}}   & 0.658{\std{±0.052}}   & 0.412{\std{±0.031}}   & 1.704{\std{±0.157}} \\
            CTA(VAE-AE)             & 0.237{\std{±0.000}}   & 0.513{\std{±0.007}}   & \cellcolor{gray!20} \textbf{0.335{\std{±0.008}}}   & \cellcolor{gray!20} \textbf{1.283{\std{±0.107}}} \\
            CTA(AE-AE)              & \cellcolor{gray!20} \textbf{0.235{\std{±0.000}}}   & \cellcolor{gray!20} \textbf{0.508{\std{±0.008}}}   & 0.364{\std{±0.009}}   & 1.365{\std{±0.074}} \\
    
    \specialrule{1pt}{1pt}{1pt}
    \end{tabular}
}
\vspace{-1em}
\end{table}

\subsection{Comparison to Interpolation method}
We use the natural cubic spline method to build $X(t)$. We report the performance of the interpolation itself in the extreme case of the 70\% missing rate in \texttt{AirQuality} and \texttt{Stocks}. As shown in Table~\ref{table:sensitivity_interpolation}, it can be observed that the performance is improved compared to the interpolation alone.

\section{Conclusion}

In this paper, we tackled how to impute regular and irregular time series. We presented a novel method based on NCDEs, which generalizes (variational) autoencoders in a continuous manner. Our method creates one (variational) autoencoder every time point and therefore, there are an infinite number of (variational) autoencoders along the time domain $[0,T]$. For this, the ELBO loss is calculated after solving an integral problem. Therefore, training occurs for every time point in the time domain, which drastically increases the training efficacy. We also presented a dual-layered architecture.

In our experiments with 4 datasets and 19 baselines, our presented method clearly marks the best accuracy in all cases. Moreover, our models have much smaller numbers of parameters than those of the state-of-the-art method. Our ablation and sensitivity studies also justify our method design. SAITS also has a dual-transformer architecture. Therefore, the main difference between our method and SAITS is that our method continuously generalizes (variational) autoencoders.

In the future, our method can be adopted to solving other time series tasks, such as classification and forecasting. Time series synthesis~\cite{yoon2019time,jeon2022gt} is also one more application to which our method can be applied.

\section*{Ethical Consideration}
Our model focuses on advancing the time series imputation. While our model itself hasn't introduced any new ethical issues, it brings to light potential concerns regarding privacy and anonymity. As data is harnessed for imputation, it's crucial to thoughtfully address the ethical considerations surrounding the confidentiality of sensitive information and the preservation of individual anonymity. balancing between data utility and safeguarding personal privacy will be crucial in ensuring the responsible and trustworthy deployment of our model. As we continue to improve and implement our model, we are committed to maintaining the highest standards of ethics and privacy, and to promoting discussions on integrating solutions that acknowledge these concerns and prioritize the well-being of all stakeholders involved.

\begin{acks}
This work was supported by Institute of Information \& communications Technology Planning \& Evaluation (IITP) grant funded by the Korea government(MSIT) (No. 2020-0-01361, Artificial Intelligence Graduate School Program at Yonsei University, 1\%), and (No. 2022-0-01032, Development of Collective Collaboration Intelligence Framework for Internet of Autonomous Things, 99\%)
\end{acks}

\bibliographystyle{ACM-Reference-Format}
\balance
\bibliography{WSDM2024}

\end{document}